\documentclass[a5paper]{llncs}
\usepackage{tabularx}
\usepackage{amssymb}

\usepackage[T1, T2A]{fontenc}
\usepackage[utf8]{inputenc}
\usepackage{indentfirst}

\usepackage[english]{babel}
\usepackage{csquotes}

\usepackage{amsmath}
\usepackage[authordate,backend=biber,dashed=false]{biblatex-chicago}

\usepackage{geometry}
\usepackage{graphicx}
\graphicspath{ {images/} }
\usepackage[dvipsnames]{xcolor}
\usepackage{ragged2e}
\usepackage{array}
\usepackage{changepage}
\usepackage{fancyhdr}
\usepackage{environ}
\usepackage[colorlinks = true,
            linkcolor = Sepia,
            urlcolor  = Mahogany,
            citecolor = BrickRed,
            anchorcolor = Sepia]{hyperref}
\usepackage{footnote}
\makesavenoteenv{tabular} 

\definecolor{light_gray}{rgb}{0.5, 0.5, 0.5}
\definecolor{dark_gray}{rgb}{0.4, 0.4, 0.4}
\definecolor{darker_gray}{rgb}{0.2, 0.2, 0.2}

\newcommand{\ddoi}[1]{\noindent\textcolor{dark_gray}{\parbox{\textwidth}{\texttt{DOI #1}}}}
\newcommand{\udc}[1]{\noindent\textcolor{dark_gray}{\parbox{\textwidth}{\texttt{UDC #1}}}} 

\renewcommand{\title}[1]{\begin{center}\Large\bfseries\boldmath#1\end{center}}
\renewcommand{\author}[1]{#1 \newline}
\newcommand{\institution}[1]{
	\textcolor{darker_gray}{\textit{\small #1}} \newline
}
\renewcommand{\email}[1]{{\textcolor{light_gray}{\small #1}} \newline}
\newcommand{\orcid}[1]{
	\textcolor{light_gray}{\texttt{\small ORCID: #1}} \newline
}

\newenvironment{main}
		{\begin{tabular}{ l @{~} @{~} !{\vrule width 2pt}| @{~} l }}
		{\end{tabular}}
\NewEnviron{abstractenv}{%
		\parbox[t]{0.6\textwidth}{	
			\begin{adjustwidth}{-1.5cm}{}	
				\begin{abstract}							
					\BODY						
				\end{abstract}
			\end{adjustwidth}
		}
}%
\NewEnviron{authorsenv}{%
  \parbox[t]{0.4\textwidth}{%
      \BODY
   }
}%

\usepackage{comment} 
\bibliography{refs} 

\begin{document}
\label{firstpage}
\sloppy 
%
%
\renewcommand{\keywords}[1]{\textbf{\noindent KEYWORDS:} \raggedright #1}
\renewcommand{\abstractname}{\noindent ABSTRACT:} 
\renewcommand\refname{References}
\renewcommand\tablename{Table}
\renewcommand\figurename{Figure}
\newcommand{\arrivaldate}[1]{{\textbf{PAPER SUBMITTED:}}{{\hfill #1}} \\}
\newcommand{\acceptdate}[1]{{\textbf{PAPER ACCEPTED:}}{{\hfill #1}}}
\newcommand{\dates}[3]{%
	\vspace{#1cm}
	\noindent\parbox[b]{0.57\textwidth}{%
		\arrivaldate{#2} 
		\acceptdate{#3} 
	}	
}%

\title{New Textual Corpora\\ for Serbian Language Modeling} 

\udc{} 
\ddoi{}  

\begin{main} 

        \begin{abstractenv}  
		This paper will present textual corpora for Serbian (and Serbo-Croatian), usable for the training of large language models and publicly available at one of the several notable online repositories. Each corpus will be classified using multiple methods and its characteristics will be detailed. Additionally, the paper will introduce three new corpora: a new umbrella web corpus of Serbo-Croatian, a new high-quality corpus based on the doctoral dissertations stored within National Repository of Doctoral Dissertations from all Universities in Serbia, and a parallel corpus of abstract translation from the same source. The uniqueness of both old and new corpora will be accessed via frequency-based stylometric methods, and the results will be briefly discussed.

		\keywords{corpora, Serbian language, language models, evaluation.} 
			
		\dates{0.4}{12 April 2024}{13 May 2024} 
		\end{abstractenv}

& 

            \begin{authorsenv}
			\author{Mihailo Škorić} 
	\email{mihailo.skoric@rgf.bg.ac.rs}  
 	\orcid{0000-0003-4811-8692}  
			\institution{University of Belgrade} 
			\institution{Faculty of Mining and Geology} 	
			\institution{Belgrade, Serbia} 	

   			\author{Nikola Janković} 
	\email{nikolajankovickv@gmail.com}  
 	\orcid{0000-0003-3484-4220}  
			\institution{University of Belgrade} 
			\institution{Faculty of Philology} 	
			\institution{Belgrade, Serbia} 	

		\end{authorsenv}

\end{main}

\section{Introduction}\label{sec:1}

With the abrupt increase of available textual data via \textit{Big Data} phenomenon near the beginning of the twenty-first century, it was soon realized that that data can be used to build corpora for natural language modeling. The fast-growing web-based data was first used as an add-on to the existing slower-growing book-based data, but with an increased interest in quantity, it slowly but surely caught up with and surpassed the latter’s share in the various language model training corpora. Today, most of the publicly available corpora use web-based data, mostly due to looser copyright constraints. 

In the context of this research, we will categorize the datasets into the following categories based on their origin:

\begin{description}
\item[$O_1$]{Web corpora – corpora containing texts collected from the open web, primarily using automatic scraping of HTML pages;}
\item[$O_2$]{Textbook corpora – corpora containing texts from textbooks or other similar origin, namely scientific publications of all sorts (scientific monographs, journal articles, conference/proceedings papers, theses etc.), and other published non-literary works (government and legal texts not present on the web, philosophy, kitchen recipes etc.) collected primarily from the digitally-born documents or physical copy scans;}
\item[$O_3$]{Literary corpora – corpora containing published literary works including mythology and religious texts;}
\item[$O_4$]{Synthetic corpora – corpora containing non-web texts created by machines using natural language generation or machine translation;}
\item[$O_5$]{Mixed corpora – corpora created using texts originating from a mix of the aforementioned and other sources.}
\end{description}

Apart from the obvious difference in source material, these classes are also differentiated by the creation process for the containing texts: while texts from the first three groups ($O_1$-$O_3$) are mostly human-created (no one can guarantee that scrapped HTML page is not machine generated text), only the second and third group contain those texts that are necessarily curated, i.e. they were read and corrected before publication, which takes time, but ensures higher quality. Texts created by machines ($O_4$) are usually prepared the fastest of all but are also associated with lesser quality.

Another important classification is corpora form. Regarding that, we also classify corpora in the following way:

\begin{description}
\item[$F_1$]{Plain-textual corpora -- corpora incorporating only plain texts which can be used to pre-train large language models such as BERT\footnote{Bidirectional Encoder Representations from Transformers}~\autocite{devlin2018bert} and GPT\footnote{Generative Pre-trained Transformer}~\autocite{
radford2018improving};}
\item[$F_2$]{Annotated corpora -- corpora where tokens, sentences or other spans are annotated, e.g. part-of-speech-annotated texts or sentences labeled with sentiment -- these corpora are usually used to fine-tune models on downstream annotation and labeling tasks;}
\item[$F_3$]{Parallel corpora -- corpora where each sentence (or other span) has a specific pair, e.g. a translation to another language or an answer to a question -- these are usually used for the training of models with a generative component such as T5\footnote{Text-To-Text Transfer Transformer} model~\autocite{raffel2020exploring}.}
\end{description}

In the following section, the paper will focus on the Serbian and Serbo-Croatian corpora procured from three notable online corpora/dataset sources: Hugging Face\footnote{\href{https://huggingface.co}{huggingface.co}, the largest web hub for publishing language models.}, CLARIN virtual language repository\footnote{\href{https://www.clarin.eu}{www.clarin.eu}, digital infrastructure offering data, tools and services to support research based on language resources.} and European Language Grid\footnote{\href{https://live.european-language-grid.eu}{live.european-language-grid.eu}, platform for all European Language Technologies originated from MetaNet.}. Other singular repositories (e.g. GitHub projects) were not surveyed.
Basic information about corpora retrieved from these hubs will be presented in Section~\ref{sec:2}, newly-prepared corpora will be presented in 
Section~\ref{sec:3}, and the experiment in accessing their uniqueness will be presented in 
Section~\ref{sec:4}. Finally, the results of the 
assessment will be discussed in Section~\ref{sec:6}.

\section{Available Corpora}
\label{sec:2}

As already mentioned, this section will present a list of corpora compiled by skimming three online sources: Hugging Face (HF), CLARIN virtual language repository (VLO) and European Language Grid (ELG) with corpora languages being limited to the domain of Serbo-Croatian macro language i.e., Serbian, Montenegrin, Croatian and Bosnian as training with closely related languages can be beneficial in certain scenarios, particularly for tasks involving multilingual understanding or translation. Retrieved corpora were categorized both by form (primary classification) and origin (secondary classification). For plain-textual corpora ($F_1$), minimum inclusion requirement was thirty million words for web corpora ($O_1$), and three million words for the others ($O_2-O_5$). Since most of the corpora in this category are web-originated, a brief study on deduplication was also conducted. As for the annotated ($F_2$) and parallel corpora ($F_3$), the size requirement was set to 3000 sentences, as these resources are much scarcer and usually used for fine-tuning only.

\subsection{Publicly available plain-text corpora}

Tables \ref{table:1} and \ref{table:2} detail the twenty-four retrieved plain-text corpora. The first table focuses on the Serbian corpora only,
while the second presents the other corpora in the scope of the Serbo-Croatian macro language, including several \textit{umbrella} corpora (produced via aggregation and deduplication of several others).

\begin{table}[h!]
\newcolumntype{Y}{>{\centering\arraybackslash}X}
\centering
\renewcommand{\arraystretch}{1.2}
\vspace{0cm}
\begin{tabularx}{\textwidth}{ | c | Y | Y | Y | c | Y | }

  \hline 
\textbf{Name} & \textbf{Lang.} &  \textbf{Origin}  &  \textbf{Size} & \textbf{Publisher} &  \textbf{Hub} \\ \hline  \hline
srWaC & sr &  web &  493  & ReLDI & \href{https://www.clarin.si/repository/xmlui/handle/11356/1063}{VLO}\\ \hline
cc100\_sr & sr &  web & 711 & Conneau et al. & \href{https://huggingface.co/datasets/cc100}{HF}\\ \hline
mC4-sr & sr &  web & 800 & Google & \href{https://huggingface.co/datasets/allenai/c4}{HF}\\ \hline
OSCAR-sr & sr &  web & 632 & OSCAR proj. & \href{https://huggingface.co/datasets/oscar-corpus/OSCAR-2301}{HF}\\ \hline
CLASSLA-sr & sr &  web &  752 & CLASSLA & \href{https://www.clarin.si/repository/xmlui/handle/11356/1426}{VLO}\\ \hline
MaCoCu-sr & sr &  web &  2,491 & Bañón et al & \href{https://www.clarin.si/repository/xmlui/handle/11356/1807}{VLO}\\ \hline
PDRS1.0 & sr &  web & 602 & ISJ & \href{https://www.clarin.si/repository/xmlui/handle/11356/1752}{VLO}\\ \hline
SrpKorNews & sr &  web &468& JeRTeh & \href{https://huggingface.co/datasets/jerteh/SrpKorNews}{HF}\\ \hline
SrpELTeC & sr &  literary & 5.3 & JeRTeh & \href{https://huggingface.co/datasets/jerteh/SrpELTeC}{HF}\\ \hline
\end{tabularx}
\vspace{0.2cm}
\caption{Exclusively Serbian plain-text corpora available on surveyed dataset hubs. Size is represented in millions  of words (M).}
\label{table:1}
\end{table}

\begin{table}[h!]
\newcolumntype{Y}{>{\centering\arraybackslash}X}
\centering
\renewcommand{\arraystretch}{1.15}
\vspace{0cm}
\begin{tabularx}{\textwidth}{ | c | Y | Y | Y | c | Y | }

  \hline 
\textbf{Name} & \textbf{Lang.} &  \textbf{Origin}  &  \textbf{Size} & \textbf{Publisher} &  \textbf{Hub} \\ \hline  \hline

meWaC & cnr &  web & 80 & ReLDI & \href{https://www.clarin.si/repository/xmlui/handle/11356/1429}{VLO}\\ \hline
hrWaC & hr &  web & 1,250 & ReLDI & \href{https://www.clarin.si/repository/xmlui/handle/11356/1064}{VLO}\\ \hline
bsWaC & bs &  web & 256 & ReLDI & \href{https://www.clarin.si/repository/xmlui/handle/11356/1062}{VLO}\\ \hline
cc100\_hr & hr &  web & 2,880 & Conneau et al. & \href{https://huggingface.co/datasets/cc100}{HF}\\ \hline
CLASSLA-hr & hr &  web & 1,341 & CLASSLA & \href{https://www.clarin.si/repository/xmlui/handle/11356/1426}{VLO}\\ \hline
CLASSLA-bs & bs &  web & 534 & CLASSLA & \href{https://www.clarin.si/repository/xmlui/handle/11356/1426}{VLO}\\ \hline
hr\_news & hr &  web & 1,433 & CLASSLA & \href{https://huggingface.co/datasets/classla/xlm-r-bertic-data}{HF}\\ \hline
MaCoCu-cnr & cnr &  web & 161 & Bañón et al & \href{https://www.clarin.si/repository/xmlui/handle/11356/1809}{VLO}\\ \hline
MaCoCu-hr & hr &  web & 2,363 & Bañón et al & \href{https://www.clarin.si/repository/xmlui/handle/11356/1806}{VLO}\\ \hline
MaCoCu-bs & bs &  web & 730 & Bañón et al & \href{https://www.clarin.si/repository/xmlui/handle/11356/1808}{VLO}\\ \hline

riznica & hr &  mixed & 87 & IHJJ & \href{https://www.clarin.si/repository/xmlui/handle/11356/1180}{VLO}\\ \hline
HPLT 1.2-sh & mixed & web & 10,030 & HPLT proj. & \href{https://huggingface.co/datasets/HPLT/hplt_monolingual_v1_2}{HF}\\ \hline

BERTić-data & mixed &  $\sim$web & 8,388 & CLASSLA & \href{https://www.clarin.si/repository/xmlui/handle/11356/1426}{VLO}\\ \hline
MaCoCu-hbs & mixed &  web & 5,490 & CLASSLA & \href{https://huggingface.co/datasets/classla/xlm-r-bertic-data}{HF}\\ \hline
XLM-R-BERTić-data & mixed &  $\sim$web & 11,539 & CLASSLA & \href{https://huggingface.co/datasets/classla/xlm-r-bertic-data}{HF}\\ \hline
\end{tabularx}
\vspace{0.2cm}
\caption{Serbo-Croatian plain-text corpora available on surveyed dataset hubs, which is not exclusively Serbian. Size is represented in millions of words (M).}
\label{table:2}
\end{table}

Excluding only four corpora, the remainder are sourced exclusively from the web. One corpus is literary, one is mixed, and the remaining two are aggregated and classified as nearly pure web corpora ($\sim$web), since they incorporate the aforementioned single mixed-origin corpora, \textit{riznica}~\autocite{cavar2012riznica}.

Most of these corpora are a product of several specific efforts. Corpora \textit{srWac}, \textit{meWac}, \textit{hrWac} and \textit{bsWac} originate from a single web scrape of top-level domains for the four languages~\autocite{ljubevsic2014bs, 11356/1429}, resulting in over two billion words. The process was repeated again later in order to produce \textit{CLASSLA-sr}, \textit{CLASSLA-hr} and \textit{CLASSLA-bs}, amassing over 2.5 billion words~\autocite{ljubevsic2021berti}.

Additional six corpora were derived from the common crawl dataset: \textit{OSCAR-sr} (632M) is the OSCAR project dataset derivative~\autocite{suarez2019asynchronous}, \textit{mC4-sr} (800M) was derived from the multilingual C4 dataset by Google~\autocite{xue-etal-2021-mt5}, \textit{HPLT 1.2-sh} (over 10 billion words) was gained from the HPLT project dataset~\autocite{aulamo2023hplt}, and finally, \textit{cc100\_sr} and \textit{cc100\_hr} (over 3.5 billion words) are the cc100 dataset derivatives~\autocite{cc100}. It should be noted that some of these efforts produced additional corpora, which were not included in this study due to not meeting the minimum size requirement of three million words.

MaCoCu project web crawling effort~\autocite{banon2022macocu, kuzman23} produced \textit{MaCoCu-sr} (2.5 billion words), \textit{MaCoCu-cnr} (151M), \textit{MaCoCu-hr} (2.2 billion words), \textit{MaCoCu-bs} (686M), and their deduplicated aggregation, \textit{MaCoCu-hbs} (5.5 billion words).

Another notable deduplication effort included four {WaC} corpora, three \textit{CLASSLA} corpora, \textit{cc100\_sr}, \textit{cc100\_hr} and \textit{riznica} corpora. This produced a 8.4 billion word aggregated corpus, published under the name \textit{BERTić-data}~\autocite{ljubevsic2021berti}.

\begin{figure}[h!]
\centering
\includegraphics[width=\textwidth]{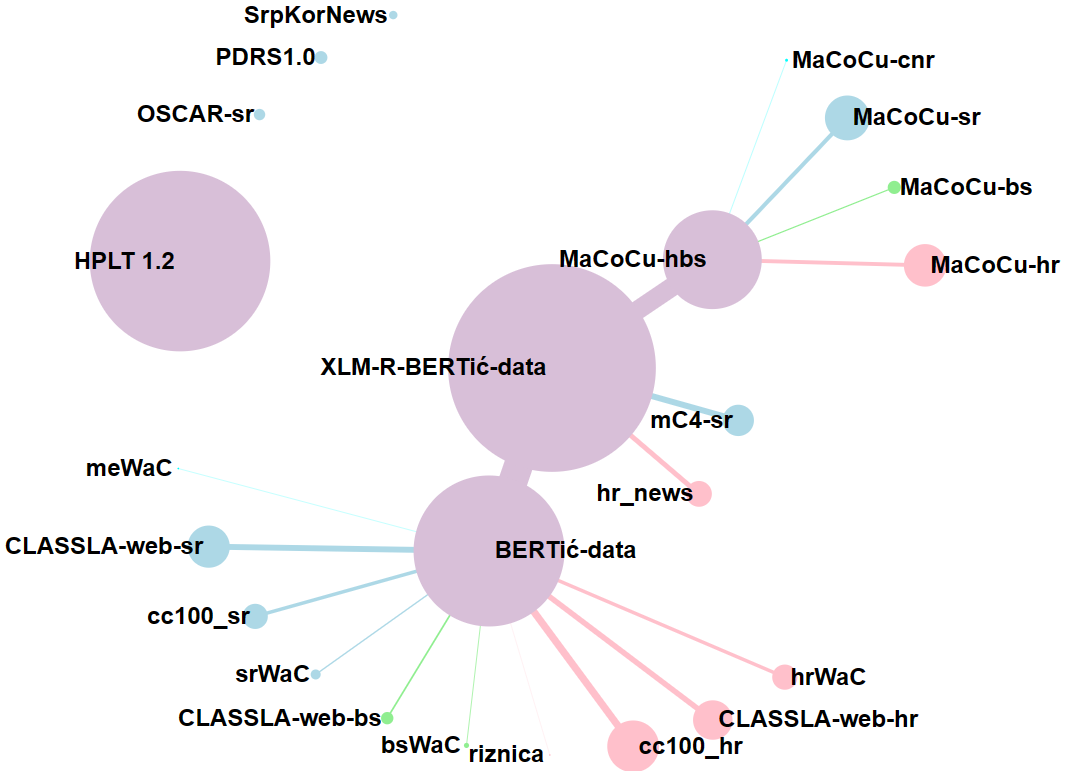}
\caption{Aggregation hierarchy of publicly available Serbo-Croatian plain-text corpora. Blue color represents Serbian corpora, cyan Montenegrin, pink Croatian, lime Bosnian and the color purple represents corpora with mixed language origin.}
\label{fig:fig1}
\end{figure}

Both \textit{BERTić-data} and \textit{MaCoCu-hbs} were aggregated again, together with \textit{mC4-sr} and \textit{hr\_news} (1.4 billion words) in order to create \textit{XLM-R-BERTić-data} (11.5 billion words), the largest published Serbo-Croatian aggregated corpus. It, however, does not incorporate common crawl incarnates \textit{HPLT 1.2-sh} and \textit{OSCAR-sr}, nor recently published \textit{PDRS1.0}~\autocite{11356/1752} (600M) nor \textit{SrpKorNews}~\autocite{CvRS2023LRS} corpora (468M). This indicates a possibility of creating new and bigger umbrella corpora for both Serbian and Serbo-Croatian. The complete current effort in corpus aggregation is visualized in Figure~\ref{fig:fig1}. 

The single literary corpus of the required size available online remains the \textit{SrpELTeC} corpus~\autocite{IJDH-CK-21,stankovic2022distant}, which includes 120 novels and boasts 5 million words. The publication process was not constrained, as the corpus comprises materials that have exceeded the threshold of copyright expiration.

\subsection{Publicly available annotated corpora}

Surveying the selected dataset hubs we found 10 annotated corpora for Serbian (Table~\ref{table:3}). We focused on the corpora annotated using the Universal POS tag set (17 classes), while there were no limitations regarding named entity recognition (NER) and sentiment annotation. In contrast to the plain-text corpora found primarily on HF and VLO, these were also found on the ELG in numbers.

POS-annotated corpora for Serbian include four resources: \textit{SrpKor4Tagging}~\autocite{stankovic-etal-2020-machine} with 60K, \textit{Intera}~\autocite{gavrilidou2004building, stankovic-etal-2020-machine} with 47.5K, \textit{1984}~\autocite{krstev2004multext} with 6.7K and \textit{reldi\_sr}~\autocite{Miličević_Ljubešić_2016} with 5.5K sentences. 
Two additional resources were annotated by both POS and NER tags: \textit{SETimes\_sr}~\autocite{samardzic-etal-2017-universal} and \textit{NormTagNER-sr}~\autocite{11356/1794} with 7K and 4K sentences respectively and 5 NER classes. 
Three additional resources were annotated via NER only: \textit{SrpELTeC-gold}~\autocite{todorovic2021serbian,frontini2020named} based on a portion of literary corpora counting 52K sentences (7 NER classes), Serbian portion of the \textit{WikiAnn} corpus~\autocite{rahimi-etal-2019-massively} with 40K sentences (3 NER classes), and Serbian portion of the \textit{polyglot\_ner} corpus~\autocite{polyglotner} with over half of a million sentences and 3 NER classes. 
The sole remaining resource is the Serbian part of the \textit{mms} sentiment-annotated corpus~\autocite{augustyniak2023massively} with 76K sentences (3 sentiment classes).

\vspace{-0.3cm}

\begin{table}[h!]
\newcolumntype{Y}{>{\centering\arraybackslash}X}
\centering
\renewcommand{\arraystretch}{1.2}
\vspace{0cm}
\begin{tabularx}{\textwidth}{ | Y | c | c | c | c | c | c |}

  \hline 
\textbf{Name} & \textbf{Lang.} &  \textbf{Origin} & \textbf{Annotat.}  &  \textbf{Size} & \textbf{Publisher} &  \textbf{Hub} \\ \hline  \hline

SrpKor4Tagging & sr &  mixed & POS & 60 & JeRTeH & \href{https://live.european-language-grid.eu/catalogue/corpus/9295}{ELG}\\ \hline
1984 & sr &  literary & POS & 6.7 & MULTEXT-East & \href{https://live.european-language-grid.eu/catalogue/corpus/8185}{ELG}\\ \hline
Intera & sr &  textbook & POS & 47.5 & META-SHARE & \href{https://live.european-language-grid.eu/catalogue/corpus/685}{ELG}\\ \hline
reldi\_sr & sr &  web &POS &  5.5 & ReLDi & \href{https://huggingface.co/datasets/classla/reldi_sr}{HF}\\ \hline
SETimes\_sr & sr &  web &POS, NER &  4 & ReLDi & \href{https://huggingface.co/datasets/classla/setimes_sr}{HF}\\ \hline
NormTagNER-sr & sr &  web &POS, NER &  7 & ReLDi & \href{https://www.clarin.si/repository/xmlui/handle/11356/1794}{VLO}\\ \hline
SrpELTeC-gold & sr &  literary &NER &  52 & JeRTeh & \href{https://live.european-language-grid.eu/catalogue/corpus/9485}{ELG}\\ \hline
wikiann-sr & sr &  $\sim$textbook & NER & 40 & Wikimedia & \href{https://huggingface.co/datasets/wikiann}{HF}\\ \hline
polyglot\_ner-sr & sr &  mixed &NER &  560 & Al-Rfou et al & \href{https://huggingface.co/datasets/polyglot_ner}{HF}\\ \hline
mms-sr & sr &  web & sentiment & 76 & Brand24 & \href{https://huggingface.co/datasets/Brand24/mms}{HF}\\ \hline

\end{tabularx}
\vspace{0.2cm}
\caption{Exclusively Serbian annotated corpora available on surveyed dataset hubs. Size is represented in thousands of sentences (K).}
\label{table:3}
\end{table}

\vspace{-0.2cm}

A similar situation was observed regarding the other Serbo-Croatian languages, where another ten resources had been retrieved (Table~\ref{table:4}), with the list mostly consisting of the Croatian counterparts of the same, previously mentioned annotated resources. These include annotated Croatian translation of the \textit{1984}, \textit{reldi\_hr}, \textit{NormTagNER-hr}, Croatian part of the \textit{WikiAnn} and \textit{polyglot\_ner} corpora and the Croatian sentiment-annotated sentences from the \textit{mms} corpora. Additionally, there is the Bosnian \textit{mms} and Serbo-Croatian \textit{WikiAnn} extracted from the Serbo-Croatian Wikipedia articles. The sizes of these corpora are mostly comparable to those of their Serbian counterparts.

The sole unique resource from this list is \textit{hr500k}~\autocite{LJUBEI16.340}, which was annotated with NER tags (5 classes) in its totality (25k sentences), and only partially with Universal POS tag set (9k sentences).

\begin{table}[h!]
\newcolumntype{Y}{>{\centering\arraybackslash}X}
\centering
\renewcommand{\arraystretch}{1.2}
\vspace{0cm}
\begin{tabularx}{\textwidth}{ | Y | c | c | c | c | c | c |}

  \hline 
\textbf{Name} & \textbf{Lang.} &  \textbf{Origin} & \textbf{Annotat.}  &  \textbf{Size} & \textbf{Publisher} &  \textbf{Hub} \\ \hline  \hline

1984 & hr &  literary & POS & 6.7 & MULTEXT-East & \href{https://live.european-language-grid.eu/catalogue/corpus/8185}{ELG}\\ \hline
hr500k & hr &  literary &POS &  9 & ReLDi & \href{https://huggingface.co/datasets/classla/hr500k}{HF}\\ \hline
reldi\_hr & hr &  web &POS &  7 & ReLDi & \href{https://huggingface.co/datasets/classla/reldi_hr}{HF}\\ \hline
NormTagNER-hr & hr &  web &POS, NER &  8 & ReLDi & \href{https://www.clarin.si/repository/xmlui/handle/11356/1793}{VLO}\\ \hline
hr500k & hr &  literary &NER &  25 & ReLDi & \href{https://huggingface.co/datasets/classla/hr500k}{HF}\\ \hline
wikiann-hr & hr &  $\sim$textbook & NER & 40 & Wikimedia & \href{https://huggingface.co/datasets/wikiann}{HF}\\ \hline
wikiann-sh & sh &  $\sim$textbook & NER & 40 & Wikimedia & \href{https://huggingface.co/datasets/wikiann}{HF}\\ \hline
polyglot\_ner-hr & hr &  mixed &NER &  630 & Al-Rfou et al & \href{https://huggingface.co/datasets/polyglot_ner}{HF}\\ \hline
mms-hr & hr &  web & sentiment & 78 & Brand24 & \href{https://huggingface.co/datasets/Brand24/mms}{HF}\\ \hline
mms-bs & bs &  web & sentiment & 36 & Brand24 & \href{https://huggingface.co/datasets/Brand24/mms}{HF}\\ \hline

\end{tabularx}
\vspace{0.3cm}
\caption{Serbo-Croatian annotated corpora available on surveyed dataset hubs, which is not exclusively Serbian. Size is represented in thousands of sentences (K).}
\label{table:4}
\end{table}

\subsection{Publicly available parallel corpora}

A total of thirteen parallel resources were found for Serbian (Table~\ref{table:5}), including five corpora of parallel translations, one corpus with text-summary pairs, one corpus of paraphrases, two corpora of question-answer pairs (QA) and four instruct datasets (textual instruction and resolution pairs).

All of the QA and instruct corpora found are synthetic in origin, created using machine translation over existing QA and instruct corpora for English: \textit{m\_mmlu-sr} and \textit{m\_hellaswag-sr} are translations of datasets \textit{mmlu}~\autocite{hendrycks2021ethics} and \textit{HellaSwag}~\autocite{zellers2019hellaswag} using GPT3.5, respectively translated; \textit{airoboros-3.0-sr} is the translation of the \textit{airoboros-3.0} dataset~\autocite{tunstall2023zephyr}, and \textit{open-orca-slim-sr}, \textit{ultrafeedback-bin-sr} and \textit{alpaca-cleaned-sr} are Serbian versions of the \textit{SlimOrca}~\autocite{SlimOrca}, \textit{UltraFeedback}~\autocite{cui2023ultrafeedback} and \textit{alpaca-cleaned}~\autocite{alpaca} datasets, respectively, translated using \textit{Google translate} service.
Together with the Serbian portion of the semi-synthetic \textit{tapaco} set~\autocite{scherrer_yves_2020_3707949} (original text may not be synthetic, but the sentences were paired using a simple algorithm), most of the available parallel resources for Serbian are synthetic.

As for the non-synthetic resources, five parallel translation corpora (English translations) include literary corpora \textit{80 jours}~\autocite{vitas2008tour} with 3.7K and \textit{biblenlp-sr} derived from the electronic version of the \textit{Bible} with 31K senteces, parllel English-Serbian version of  \textit{Intera} (47.5K sentences), Serbian-English parallel subset of the \textit{Opus} dataset~\autocite{tiedemann-2012-parallel} with 1 million sentences, and the parallelized subset of the \textit{MaCoCu} set, \textit{MacocuParallel-sr}~\autocite{banon2022macocu} with 2 million sentences. While the two web-originated corpora are much more massive in size, they may or may not need to be deduplicated.

The sole summary dataset is \textit{XL-Sum} with 15K summarized text pairs~\autocite{hasan-etal-2021-xl}.

\begin{table}[h!]
\newcolumntype{Y}{>{\centering\arraybackslash}X}
\centering
\renewcommand{\arraystretch}{1.2}
\vspace{0cm}
\begin{tabularx}{\textwidth}{ | Y | c | c | c | c | c | c |}

  \hline 
\textbf{Name} & \textbf{Lang.} &  \textbf{Origin} & \textbf{Pair type}  &  \textbf{Size} & \textbf{Publisher} &  \textbf{Hub} \\ \hline  \hline

80 jours & sr &  literary & translation &  3.7 & JeRTeh & \href{https://live.european-language-grid.eu/catalogue/corpus/13141}{ELG}\\ \hline

biblenlp-sr & sr &  literary & translation &  31 & eBible & \href{https://huggingface.co/datasets/bible-nlp/biblenlp-corpus}{HF}\\ \hline

Intera & sr &  mixed & translation &  47.5 & META-SHARE & \href{https://live.european-language-grid.eu/catalogue/corpus/657}{ELG}\\ \hline

OPUS-sr & sr &  web & translation & 1000 & Opus project & \href{https://huggingface.co/datasets/Helsinki-NLP/opus-100}{HF}\\ \hline
MacocuParallel-sr & sr &  web & translation & 2000 & Bañón et al & \href{https://huggingface.co/datasets/MaCoCu/parallel_data}{HF}\\ \hline

XL-Sum & sr &  web & summary & 15  & Hasan et al & \href{https://live.european-language-grid.eu/catalogue/corpus/9110}{ELG}\\ \hline

tapaco-sr & sr &  $\sim$synthetic & paraphrase &  8 & Scherrer, Yves & \href{https://huggingface.co/datasets/tapaco}{HF}\\ \hline
m\_mmlu-sr & sr &  synthetic & QA & 14.5 & Alexandra Inst. & \href{https://huggingface.co/datasets/alexandrainst/m_mmlu}{HF}\\ \hline
m\_hellaswag-sr & sr &  synthetic &QA &  9.5 & Alexandra Inst. & \href{https://huggingface.co/datasets/alexandrainst/m_hellaswag}{HF}\\ \hline
airoboros-3.0-sr & sr &  synthetic & instruction & 46 & draganjovanovich & \href{https://huggingface.co/datasets/draganjovanovich/airoboros-3.0-serbian}{HF}\\ \hline
open-orca-slim-sr & sr &  synthetic & instruction & 515 & DataTab & \href{https://huggingface.co/datasets/datatab/open-orca-slim-serbian}{HF}\\ \hline
ultrafeedback-bin-sr & sr &  synthetic & instruction & 63 & DataTab & \href{https://huggingface.co/datasets/datatab/ultrafeedback_binarized_serbian}{HF}\\ \hline
alpaca-cleaned-sr & sr &  synthetic & instruction & 52 & DataTab & \href{https://huggingface.co/datasets/datatab/alpaca-cleaned-serbian-full}{HF}\\ \hline
\end{tabularx}
\vspace{0.3cm}
\caption{Exclusively Serbian parallel corpora available on surveyed dataset hubs. Size is represented in thousands of pairs (K).}
\label{table:5}
\end{table}

Non-Serbian parallel sets are presented in Table~\ref{table:6} and consist mostly of the alternative subsets of the same resources: English-Croatian subset of \textit{biblenlp}, subsets of \textit{OPUS} dataset containing translations of Serbo-Croatian, Croatian and Bosnian to English, and subsets of the \textit{MaCoCu Parallel} dataset containing Translations of Montenegrin, Croatian and Bosnian into English. These make most of the sentences, counting over 5 million.

Two unique resources from the list are the Croatian subset of the \textit{mfaq} dataset made up out of 5k web-scraped QA pairs~\autocite{debruyn2021mfaq} and the Serbo-Croatian subset of the \textit{exams} dataset containing 5k QA pairs taken from actual student tests~\autocite{hardalov-etal-2020-exams}. Despite the relatively small size, these resources are important in the heavily underrepresented group of QA datasets, the only alternative being synthetic translations.

\begin{table}[h!]
\newcolumntype{Y}{>{\centering\arraybackslash}X}
\centering
\renewcommand{\arraystretch}{1.2}
\vspace{0cm}
\begin{tabularx}{\textwidth}{ | Y | c | c | c | c | c | c |}

  \hline 
\textbf{Name} & \textbf{Lang.} &  \textbf{Origin} & \textbf{Pair type}  &  \textbf{Size} & \textbf{Publisher} &  \textbf{Hub} \\ \hline  \hline

biblenlp-hr & hr &  literary & translation &  31 & eBible & \href{https://huggingface.co/datasets/bible-nlp/biblenlp-corpus}{HF}\\ \hline

OPUS-sh & sh &  web & translation & 271 & Opus project & \href{https://huggingface.co/datasets/Helsinki-NLP/opus-100}{HF}\\ \hline
OPUS-hr & hr &  web & translation & 1000 & Opus project & \href{https://huggingface.co/datasets/Helsinki-NLP/opus-100}{HF}\\ \hline
OPUS-bs & bs &  web & translation & 1000 & Opus project & \href{https://huggingface.co/datasets/Helsinki-NLP/opus-100}{HF}\\ \hline

MacocuParallel-me & cnr &  web & translation & 218 & Bañón et al & \href{https://huggingface.co/datasets/MaCoCu/parallel_data}{HF}\\ \hline
MacocuParallel-hr & hr &  web & translation & 2000 & Bañón et al & \href{https://huggingface.co/datasets/MaCoCu/parallel_data}{HF}\\ \hline
MacocuParallel-bs & bs &  web & translation & 500 & Bañón et al & \href{https://huggingface.co/datasets/MaCoCu/parallel_data}{HF}\\ \hline

mfaq-hr & hr &  web & QA &  5 & CLiPS & \href{https://huggingface.co/datasets/clips/mfaq}{HF}\\ \hline
exams-sh & mixed &  textbook & QA &  5 & Hardalov et al & \href{https://huggingface.co/datasets/mhardalov/exams}{HF}\\ \hline

\end{tabularx}
\vspace{0.3cm}
\caption{Serbo-Croatian parallel corpora available on surveyed dataset hubs, which is not exclusively Serbian. Size is represented in thousands of pairs (K).}
\label{table:6}
\end{table}

\vspace{-0.4cm}

\section{New Corpora}
\label{sec:3}

A total of three new corpora are envisioned in the scope of this research. The first (\textit{Umbrella corp}) is the new and larger aggregated (umbrella) plain-text corpus that will encompass the entirety of currently published web corpora under a sole entity. The creation of \textit{Umbrella corp} is elaborated in Section~\ref{sec:3.1}. The second envisioned corpus (\textit{S.T.A.R.S.}) is designed to boost the representation of openly available textbook-quality sources and it is based on the doctoral dissertations downloaded from the NaRDuS platform\footnote{\href{https://nardus.mpn.gov.rs/} {NaRDuS} -- National Repository of Doctoral Dissertations from all Universities in Serbia.}, which were previously used to train currently largest language model pre-trained for Serbian, \textit{gpt2-orao}~\autocite{skoric24modeli}. The preparation process for \textit{S.T.A.R.S.} will be explained in more detail in Section~\ref{sec:3.2}. The final corpus (\textit{PaSaž}) represents the aligned translations of abstracts extracted from those dissertations using methods presented in Section~\ref{sec:3.3}.

\subsection{Aggregated web corpus - \textit{Umbrella corp}}
\label{sec:3.1}
New umbrella Serbo-Croatian web corpus, \textit{Umbrella corp.} is an aggregation of the twenty existing corpora surveyed and discussed in this paper. We avoided using existing aggregated corpora as material to ensure the possibility of specific language extraction later on, e.g. the Serbian portion of the corpora, but it should be noted that the most of the used corpora had already been deduplicated both against other corpora from the list and internally. The only mixed-language corpora on the list, \textit{HPLT 1.2-sh} was additionally processed to split it into corpora containing documents in Serbian (\textit{HPLT-sr})and Croatian i.e. non Serbian (\textit{HPLT-hr}). The basis for this fuzzy classification was the frequency of specific words (\textit{tko/ko}, \textit{što/šta} ,\textit{uvjet/uslov}, \textit{uopće/uopšte} etc.) and the ratio of the character \textit{e} and substring \textit{je} indicating Ijekavian dialect.

The corpora was additionally cleaned and deduplicated. Deduplication of the corpora content was performed on all documents using \textit{onion}~\autocite{pomikalek2011removing} corpus processing tool which performs fuzzy deduplication, and locates duplicate documents on the bases of n-grams duplicated across the whole corpus. For this experiment we used \textit{6-gram} duplicates search and we eliminated all documents over 75\% n-gram duplication threshold.

After this deduplication the the total count of words was reduced by a third, from 28 billion words to 18.6 billion.  With this total count, \textit{Umbrella corp} becomes the biggest available umbrella plain-text corpus in the language scope, with the additional improvements (e.g. additional boilerplate removal and text correction) planned in the future.

The complete list of corpora used, their sizes in millions of words before and after deduplication and cleaning, as well as their total share in the final version of the \textit{Umbrella corp.} is presented in Table~\ref{table:x}.

\begin{table}[h!]
\newcolumntype{Y}{>{\centering\arraybackslash}X}
\centering
\renewcommand{\arraystretch}{1.2}
\vspace{0cm}
\begin{tabularx}{\textwidth}{ | Y | Y | Y | Y | Y | }

  \hline 
Name & Language & Size before & Size after & Share \\ \hline \hline
srWaC & Serbian &  493 &  307  &  1.65\% \\ \hline
meWaC & Montenegrin &  80 & 41 &  0.22\%\\ \hline
hrWaC & Croatian &  1250 & 935 & 5.01\% \\ \hline
bsWaC & Bosnian &  256 & 194 & 1.04\%  \\ \hline

OSCAR-sr & Serbian &  632 & 410 &  2.20\% \\ \hline

cc100\_hr & Croatian &  2,880 & 2,561 & 13.73\% \\ \hline
cc100\_sr & Serbian &  711 & 659 & 3.53\% \\ \hline

CLASSLA-sr & Serbian &  752 &  240 & 1.29\% \\ \hline
CLASSLA-hr & Croatian &  1341 & 160 &  0.86\%\\ \hline
CLASSLA-bs & Bosnian &  534 & 105 &  0.56\%\\ \hline

hr\_news & Croatian &  1,433 & 1,426 & 7.65\% \\ \hline
mC4-sr & Serbian &  800 & 782 & 4.19\% \\ \hline

MaCoCu-sr & Serbian &  2,491 &  2,152 &  11.54\%\\ \hline
MaCoCu-cnr & Montenegrin &  161 & 152 & 0.82\% \\ \hline
MaCoCu-hr & Croatian &  2,363 & 2,355 & 12.63\% \\ \hline
MaCoCu-bs & Bosnian &  730 & 700 & 3.75\% \\ \hline

riznica & Croatian &  87 & 69 & 0.37\% \\ \hline
PDRS1.0 & Serbian &  602 & 506 & 2.71\% \\ \hline
SrpKorNews & Serbian &  469 & 469 & 2.51\% \\ \hline

HPLT-sr & Serbian & $\sim$5,015 & 2,562 & 9.95\% \\ \hline \hline
HPLT-hr & Croatian & $\sim$5,015 & 1,856 & 13.74\% \\ \hline \hline

Total & & 28,095 &\textbf{ 18,641 }& 100\% \\ \hline
\end{tabularx}
\vspace{0.2cm}
\caption{List of corpora used to build new umbrella corpus, their sizes before and after cleaning and deduplication and their share in the final corpus. Sizes are represented in millions  of words (M).}
\label{table:x}
\end{table}

\subsection{Set of theses and academic research in Serbian - \textit{S.T.A.R.S.} }
\label{sec:3.2}

In the domain of textbook corpora (O\textsubscript{2}), doctoral theses represent a highly desirable resource type due to several factors. Firstly, the high academic standards regarding the methodology, data, and linguistic quality that a doctoral thesis needs to satisfy, as well as the peer review process that it undergoes, ensure a reliable high quality of the data from this source type. Secondly, since a doctoral dissertation is expected to be a unique scientific contribution that would advance the relevant science field further, a corpus of doctoral dissertations is uniquely positioned in terms of the thematic variety across various domains of knowledge and the timeliness of the topics described. Finally, the large number of open-access doctoral dissertations in the NaRDuS repository, the fact that certain sections of doctoral dissertations are standardized, as well as the availability of structured metadata for each dissertation on NaRDuS, make this resource the ideal candidate for a unique, large, and high-quality scientific corpus in Serbian. 

NaRDuS was envisioned within the structural TEMPUS project 5440932013, RODOS\footnote{\href{http://rodos.edu.rs/}{rodos.edu.rs} Restructuring of Doctoral Studies in Serbia (RODOS)}. According to the amendments to the Law on Higher Education\footnote{\href{https://www.pravno-informacioni-sistem.rs/SlGlasnikPortal/reg/viewAct/5f688c8e-4798-470f-9adf-5bc0739c72aa}{www.pravno-informacioni-sistem.rs/SlGlasnikPortal/reg/viewAct/5f688c8e-4798-470f-9adf-5bc0739c72aa}} (Amendments published in the Službeni glasnik Gazette, no. 99/2014, entered into force on 19.9.2014), all higher education institutions have an obligation to make PhD dissertations available to the public prior to their defense, while universities have an obligation to provide an open access digital repository of the defended dissertations. In accordance with the above, the NaRDuS platform has been in operation since September 2015, and it is based on the DSpace software, which supports the OAI-PMH protocol for metadata transfer~\autocite{verbic__2017}. At the moment of writing this article, there are 13,289 doctoral dissertations on NaRDuS.

As the first step in creating the corpus, the full list of dissertations was obtained from the sitemap of the NaRDuS website\footnote{\href{http://nardus.mpn.gov.rs/htmlmap}{nardus.mpn.gov.rs/htmlmap}}, with Python’s \textit{Requests} module, complying with the site’s robots.txt page. \textit{Selenium}
and Python were used to retrieve the full detailed metadata of each dissertation, including download links. Metadata for each dissertation was also enriched with four additional fields:

\begin{description}
    \item [\textit{fulltext\_url}] - In the cases where multiple download links were provided for one dissertation (which often included the committee report), all of the PDF documents from the links were downloaded, their page count and number of lines were automatically retrieved using the \textit{PyMuPDF} module in Python, and the appropriate link was selected and stored in this filed;

    \item [\textit{need\_ocr}] - Automatic text retrieval using \textit{PyMuPDF} for the first 10 pages was attempted to evaluate whether the documents from the selected URLs are digital-born or they would require further steps of optical character recognition (OCR) and manual revision, and the Boolean results of the procedure were stored in this field;

    \item [\textit{srpski}] - This field indicates whether the dissertations were written in Serbian (\textit{yes} value) or another language (\textit{no} value), and the language of each text was determined through examination of both \textit{dc.language} and \textit{dc.language.iso} fields;

    \item [\textit{ARR}] - This field indicates whether the dissertations were published under copyrighted license - \textit{all rights reserved} (ARR), which was determined by examining the \textit{dc.rights.license} metadata filed.
    
\end{description}

These four fields were used to filter out proper candidate-dissertation for this rendition of the corpus. We selected the ones where the proper PDF could be obtained through the \textit{fulltext\_url}, that were written in Serbian (\textit{srpski=yes}), that do not need OCR (\textit{need\_ocr=no}), and that were not published under the \textit{all rights reserved} licence (\textit{ARR=no}). This filtering was done to extract only Serbian texts and to ensure compliance with the copyright licenses.

After applying this criterion, 11,624 dissertations remained, which is around 87.5\% of all of the dissertations in the NARDUS repository. 

Finally, the full text from each of these dissertations was extracted using \textit{PyMuPDF} and saved as TXT files which were used to build the corpus. Only the lines that are part of textual paragraphs were preserved during the compilation of the final corpus file, resulting in an over 560M words corpus.

\subsection{Parallel abstracts corpus - \textit{PaSaž}}
\label{sec:3.3}
Due to the fact that parallel language resources, especially for less widely spoken languages (like Serbian), are relatively rare, parallel corpora are smaller and fewer in number compared to their monolingual counterparts. Since abstracts of the Serbian doctoral dissertations are written in Serbian and English, they represent a valuable large, high-quality resource for the creation of a parallel Serbian-English corpus of scientific language.
However, the title, format and location of the abstracts within the document  varies significantly, depending on the relevant scientific institution. In this study, a manual analysis of a large number of dissertations from different institutions was first performed, after which the documents were separated into two groups; in the first, there were dissertations whose abstracts appeared in the text itself (either at the beginning or the end of the document), and in the second, dissertations whose abstracts were presented within a table in the \textit{Key Word Documentation} section. Out of the 11,624 dissertations, 2,594 were in the second, table-based group. Three types of data were extracted from the documents (in both Serbian and English): the abstracts, keywords, and the scientific field of the dissertation. Since the keywords are already present in the metadata, their extraction was only done for the first group of documents, where there were two sets of keywords (in case of possible differences between the two versions). For the second group, only abstracts and the scientific field of the dissertation were extracted.

Since the metadata for most of the dissertations in NaRDuS contain partial abstracts in both languages,  Python script was used to search for the beginning of each abstract (first 6 words) in the first group of documents. If the dissertation did not have a partial abstract in the metadata, or the search was unsuccessful, finding the start of the abstract was attempted using regular expressions which matched the heading string of the abstract section. Since the information in this section always appears in the order 1) abstract, 2) keywords, 3) scientific field (for both languages), an approach using regular expressions was used to identify the end of each section and the beginning of the next, after which the existing metadata for the dissertation was updated with this information.

For the second group, in which the desired sections were present in the \textit{Key Word Documentation} section, the same approach was done, using different regular expressions for the table sections. 

After these steps, 7687 parallel abstracts were extracted, and the metadata of these dissertation was updated with the abstracts and the scientific field of the dissertation. Where possible, the keywords from the dissertation texts were also extracted and stored in a new filed in the metadata (\textit{keywords\_from\_text}), in case the values differ from the ones already existing in the metadata obtained from NaRDuS.

\section{Evaluation}
\label{sec:4}

Dataset evaluation in the scope of this paper focuses on the plain-text and Serbian corpora only, and the evaluation is performed through the rough assessment of the uniqueness of each corpus i.e. how much it differs from the others. In order to assess the uniqueness aspect, we decided to conduct a word-frequency-based evaluation, inspired by an existing experiment~\autocite{rayson2000comparing}. For each candidate corpus, a single million-words excerpt was used to compile word frequencies, and then the 1000 most frequent words from each corpus excerpt were extracted. These most frequent words from each of the ten corpora excerpts were used to build a features list (a unique set of words), resulting in a total of 3,257 features from ten corpora. The relative frequencies (per million) of each feature word in each corpus were used to populate feature vectors, if the word is one of the chosen 1000 for that corpus that is, otherwise, the relative frequency value was set to 0. Once the feature vectors were populated, we calculated cosine similarities (to the power of ten) between each two to generate a corpora similarity matrix. The corpus which has the lowest relative similarity with the other is speculated to be the most unique. Calculated corpus similarity matrix is presented in Table~\ref{table:8}.

\begin{table}[h!]
\newcolumntype{Y}{>{\centering\arraybackslash}X}
\centering
\renewcommand{\arraystretch}{1.2}
\vspace{0cm}
\begin{tabularx}{\textwidth}{ | c | Y | Y | Y | Y | Y | Y | Y | Y | Y | Y | }
 \hline 
 & {\rotatebox[origin=c]{90}{srWaC}}  &   {\rotatebox[origin=c]{90}{cc100\_sr}} &  {\rotatebox[origin=c]{90}{mC4\_sr}} &   {\rotatebox[origin=c]{90}{OSCAR-sr}}  &   { \rotatebox[origin=c]{90}{ CLASSLA-sr } } &  {\rotatebox[origin=c]{90}{MaCoCu-sr}} &  {\rotatebox[origin=c]{90}{PDRS1.0}} &  {\rotatebox[origin=c]{90}{SrpKorNews}}  &  {\rotatebox[origin=c]{90}{SrpELTeC}} &  {\rotatebox[origin=c]{90}{S.T.A.R.S.}}  \\ \hline  

srWaC &  & 0.93 & 0.88 & 0.95 & 0.98 & 0.79 & 0.72 & 0.87 & 0.36 & 0.71 \\ \hline
cc100\_sr & 0.93 &  & 0.91 & 0.91 & 0.90 & 0.92 & 0.75 & 0.93 & 0.44 & 0.62 \\ \hline
mC4\_sr & 0.88 & 0.91 &  & 0.84 & 0.87 & 0.84 & 0.78 & 0.88 & 0.50 & 0.61 \\ \hline
OSCAR-sr & 0.95 & 0.91 & 0.84 &  & 0.94 & 0.78 & 0.70 & 0.82 & 0.37 & 0.69 \\ \hline
CLASSLA-sr & 0.98 & 0.90 & 0.87 & 0.94 &  & 0.75 & 0.71 & 0.85 & 0.34 & 0.70 \\ \hline
MaCoCu-sr & 0.79 & 0.92 & 0.84 & 0.78 & 0.75 &  & 0.71 & 0.89 & 0.48 & 0.52 \\ \hline
PDRS1.0 & 0.72 & 0.75 & 0.78 & 0.70 & 0.71 & 0.71 &  & 0.73 & 0.47 & 0.47 \\ \hline
SrpKorNews & 0.87 & 0.93 & 0.88 & 0.82 & 0.85 & 0.89 & 0.73 &  & 0.42 & 0.56 \\ \hline
SrpELTeC & 0.36 & 0.44 & 0.50 & 0.37 & 0.34 & 0.48 & 0.47 & 0.42 &  & 0.22 \\ \hline
S.T.A.R.S. & 0.71 & 0.62 & 0.61 & 0.69 & 0.70 & 0.52 & 0.47 & 0.56 & 0.22 &  \\ \hline\hline
\textit{Average} & 0.80 & 0.81 & 0.79 & 0.78 & 0.78 & 0.74 & 0.67 & 0.77 & 0.40 & 0.57 \\ \hline

\end{tabularx}
\vspace{0.2cm}
\caption{Word frequency-based similarity matrix of Serbian plain-text corpora available on surveyed dataset hubs. The values represent cosine similarities to the power of ten, obtained by comparing feature (word) frequency vectors.}
\label{table:8}
\end{table}

From the presented results, it is apparent that the most unique corpus according to word frequencies is \textit{SrpELTeC}, with an average similarity of 0.40, especially when it is compared to total average similarity of 0.71. It is also placed furthest from every individual corpus in the embedding matrix. 

\begin{figure}[h!]
\centering
\includegraphics[width=\textwidth]{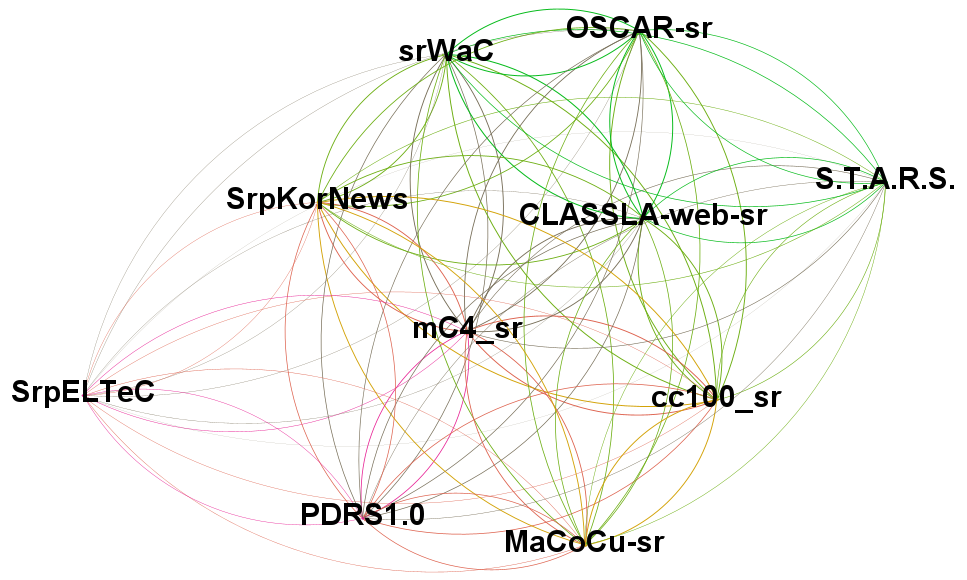}
\caption{Corpus uniqueness visualized through a two-dimensional network graph representation of the calculated corpus-similarity matrix, where the edges represent distances between each corpus and the colors represent a clustering spectrum.}
\label{fig:last}
\end{figure}

The new corpus, \textit{S.T.A.R.S.}, is the second furthest corpus from each of the other ones, and on average, making it the second most unique one (average similarity of 0.57). What is especially interesting is that the two corpora that have the lowest mutual similarity are actually \textit{S.T.A.R.S.} and \textit{SrpELTeC} with a similarity of only 0.22, making them the two opposite ends of the spectrum, with the web corpora clustering in the middle, which is particularly visible on the two-dimensional representation of the embedding matrix (Figure~\ref{fig:last}).

\section{Conclusion}\label{sec:6}
This paper provides an overview of textual corpora for Serbian (and Serbo-Croatian) language publicly available (on Hugging Face, European Language Grid and CLARIN VLO) in the following categories:

\begin{itemize}
    \item Plain text corpora - mostly web-originated, with an exception of the literary \textit{SrpELTeC} corpus;
    \item Annotated corpora - various origins, primarily annotated with POS and NER and lesser with sentiment;
    \item Parallel corpora - mostly web (\textit{MacocuParallel}, \textit{OPUS}) and literary translations and synthetic QA and instruction sets, one web summary corpus, one semi-synthetic paraphrase corpus and two smaller curated QA sets. 
\end{itemize}

We found that the vast majority of the available plain-text corpora originates from the web, and that deduplication efforts exist, but none currently encompasses all the corpora. We also find the textbook corpora highly underrepresented, with not a single corpus available matching the size criterion of three million words minimum. 

To deal with the scarcity of textbook-quality texts we propose and compile two new highly curated corpora based on publicly-available doctoral dissertations in Serbian: Plain text corpus dubbed \textit{S.T.A.R.S.} (Set of theses and academic research in Serbian) and Parallel English-Serbian corpus of abstracts dubbed \textit{PaSaž}. The former contains 11,624 documents and counts over 560 million words, and the latter contains 7,678 parallel abstracts, and additional 2,805 partial ones, counting around eight million words, a huge contribution to the publicly available textbook-quality corpora for Serbian language modeling.

We also performed a word-frequency-based evaluation that indicates the uniqueness of the dissertations in the current paradigm, where it was shown that it has relatively low similarity to other currently available corpora, especially the literary corpus \textit{SrpELTeC}, with the web corpora (having high mutual similarity) sitting in the middle (Figure~\ref{fig:last}).

Additionally, the paper introduces a new umbrella web corpus for Serbian and Serbo-Croatian dubbed \textit{Umbrella corp.} which accumulates all the currently available plain text web corpora in the language scope.

The future work in the matter should include further procurement of textbook and literary texts that can be published, and the creation of procedures for the enhancement of the existing resources, particularly web text.

\section*{Acknowledgment}

Computer resources necessary for the deduplication of the \textit{Umbrella corp.} were provided by the National Platform for Artificial Intelligence of Serbia.

This research was supported by the Science Fund of the Republic of Serbia, \#7276, \textit{Text Embeddings - Serbian Language Applications} - \textit{TESLA}.

\nocite{*}
\printbibliography

@Book{LWP-Serbian,
author = {Duško Vitas and Ljubomir Popović and Cvetana Krstev and Ivan Obradović and Gordana Pavlović-Lažetić and Mladen Stanojević},
title = {{Српски језик у дигиталном добу -- The Serbian Language in the Digital Age}},
publisher = {Springer},
year = {2012},
series = {META-NET White Paper Series. Georg Rehm and Hans Uszkoreit (Series Editors)},
isbn = {978-3-642-30754-6},
note = {Available online at \url{http://www.meta-net.eu/whitepapers}}
}

@article{IJDH-CK-21,
author = {Krstev, Cvetana},
title = {The Serbian Part of the ELTeC Collection Through the Magnifying Glass of Metadata},
journal = {Infotheca - Journal for Digital Humanities},
volume = {21},
number = {2},
year = {2021},
issn = {2217-9461},
pages = {26--42},
doi = {10.18485/infotheca.2021.21.2.2},
url = {https://infoteka.bg.ac.rs/ojs/index.php/Infoteka/article/view/2021.21.2.2_en}
}

@inproceedings{todorovic2021serbian,
  title={Serbian ner\&beyond: The archaic and the modern intertwinned},
  author={Todorovi{\'c}, Branislava {\v{S}}andrih and Krstev, Cvetana and Stankovi{\'c}, Ranka and Ne{\v{s}}i{\'c}, Milica Ikoni{\'c}},
  booktitle={Proceedings of the International Conference on Recent Advances in Natural Language Processing (RANLP 2021)},
  pages={1252--1260},
  year={2021}
}

@article{devlin2018bert,
	title        = {Bert: Pre-training of deep bidirectional transformers for language understanding},
	author       = {Devlin, Jacob and Chang, Ming-Wei and Lee, Kenton and Toutanova, Kristina},
	year         = 2018,
	journal      = {arXiv preprint arXiv:1810.04805}
}

@article{radford2018improving,
	title        = {Improving language understanding by generative pre-training},
	author       = {Radford, Alec and Narasimhan, Karthik and Salimans, Tim and Sutskever, Ilya and others},
	year         = 2018,
	publisher    = {OpenAI}
}

@article{raffel2020exploring,
	title        = {Exploring the limits of transfer learning with a unified text-to-text transformer},
	author       = {Raffel, Colin and Shazeer, Noam and Roberts, Adam and Lee, Katherine and Narang, Sharan and Matena, Michael and Zhou, Yanqi and Li, Wei and Liu, Peter J},
	year         = 2020,
	journal      = {The Journal of Machine Learning Research},
	publisher    = {JMLRORG},
	volume       = 21,
	number       = 1,
	pages        = {5485--5551}
}

@article{cavar2012riznica,
  title={Riznica: the Croatian language corpus},
  author={{\'C}avar, Damir and Brozovi{\'c} Ron{\v{c}}evi{\'c}, Dunja},
  journal={Prace filologiczne},
  volume={63},
  pages={51--65},
  year={2012}
}

@InProceedings{ljubevsic2014bs,
  author    = {Ljube{\v{s}}i{\'c}, Nikola and Klubi{\v{c}}ka, Filip},
  booktitle = {Proceedings of the 9th web as corpus workshop (WaC-9)},
  title     = {$\{$bs, hr, sr$\}$ wac-web corpora of Bosnian, Croatian and Serbian},
  year      = {2014},
  pages     = {29--35},
}

@misc{11356/1429,
 title = {Montenegrin web corpus {meWaC} 1.0},
 author = {Ljube{\v s}i{\'c}, Nikola and Erjavec, Toma{\v z}},
 url = {http://hdl.handle.net/11356/1429},
 note = {Slovenian language resource repository {CLARIN}.{SI}},
 copyright = {Creative Commons - Attribution-{ShareAlike} 4.0 International ({CC} {BY}-{SA} 4.0)},
 issn = {2820-4042},
 year = {2021} }

@misc{kuzman23,
 title = {Montenegrin web corpus {meWaC} 1.0},
 author = {Kuzman, Taja and Ljube\v{s}i\'{c}, Nikola},
 url = {https://www.clarin.si/info/k-centre/classla-web-bigger-and-better-web-corpora-for-croatian-serbian-and-slovenian-on-clarin-si-concordancers/},
 note = {CLASSLA-web: Bigger and better web corpora for Croatian, Serbian 
and Slovenian},
 year = {2023} }

@inproceedings{suarez2019asynchronous,
	title        = {Asynchronous pipeline for processing huge corpora on medium to low resource infrastructures},
	author       = {Su{\'a}rez, Pedro Javier Ortiz and Sagot, Beno{\^i}t and Romary, Laurent},
	year         = 2019,
	booktitle    = {7th Workshop on the Challenges in the Management of Large Corpora (CMLC-7)},
	organization = {Leibniz-Institut f{\"u}r Deutsche Sprache}
}

@inproceedings{xue-etal-2021-mt5,
    title = "m{T}5: A Massively Multilingual Pre-trained Text-to-Text Transformer",
    author = "Xue, Linting  and
      Constant, Noah  and
      Roberts, Adam  and
      Kale, Mihir  and
      Al-Rfou, Rami  and
      Siddhant, Aditya  and
      Barua, Aditya  and
      Raffel, Colin",
    editor = "Toutanova, Kristina  and
      Rumshisky, Anna  and
      Zettlemoyer, Luke  and
      Hakkani-Tur, Dilek  and
      Beltagy, Iz  and
      Bethard, Steven  and
      Cotterell, Ryan  and
      Chakraborty, Tanmoy  and
      Zhou, Yichao",
    booktitle = "Proceedings of the 2021 Conference of the North American Chapter of the Association for Computational Linguistics: Human Language Technologies",
    month = jun,
    year = "2021",
    address = "Online",
    publisher = "Association for Computational Linguistics",
    url = "https://aclanthology.org/2021.naacl-main.41",
    doi = "10.18653/v1/2021.naacl-main.41",
    pages = "483--498",
}

@inproceedings{aulamo2023hplt,
  title={HPLT: High Performance Language Technologies},
  author={Aulamo, Mikko and Bogoychev, Nikolay and Ji, Shaoxiong and Nail, Graeme and Ram{\'\i}rez-S{\'a}nchez, Gema and Tiedemann, J{\"o}rg and Van Der Linde, Jelmer and Zaragoza, Jaume},
  booktitle={Proceedings of the 24th Annual Conference of the European Association for Machine Translation},
  pages={517--518},
  year={2023}
}

@inproceedings{cc100,
  title={Unsupervised cross-lingual representation learning at scale},
  author={Conneau, Alexis and Khandelwal, Kartikay and Goyal, Naman and Chaudhary, Vishrav and Wenzek, Guillaume and Guzm{\'a}n, Francisco and Grave, Edouard and Ott, Myle and Zettlemoyer, Luke and Stoyanov, Veselin},
  journal={arXiv preprint arXiv:1911.02116},
  year={2019}
}

@inproceedings{banon2022macocu,
  title={MaCoCu: Massive collection and curation of monolingual and bilingual data: focus on under-resourced languages},
  author={Ban{\'o}n, Marta and Espla-Gomis, Miquel and Forcada, Mikel L and Garc{\'\i}a-Romero, Cristian and Kuzman, Taja and Ljube{\v{s}}i{\'c}, Nikola and van Noord, Rik and Sempere, Leopoldo Pla and Ram{\'\i}rez-S{\'a}nchez, Gema and Rupnik, Peter and others},
  booktitle={23rd Annual Conference of the European Association for Machine Translation, EAMT 2022},
  pages={303--304},
  year={2022},
  organization={European Association for Machine Translation}
}

@article{ljubevsic2021berti,
	title        = {BERTi{\'c}--The Transformer Language Model for Bosnian, Croatian, Montenegrin and Serbian},
	author       = {Ljube{\v{s}}i{\'c}, Nikola and Lauc, Davor},
	year         = 2021,
	journal      = {arXiv preprint arXiv:2104.09243}
}

@misc{11356/1752,
	title        = {Serbian Web Corpus {PDRS} 1.0},
	author       = {Wasserscheidt, Philipp},
	year         = 2023,
	issn         = {2820-4042},
	url          = {http://hdl.handle.net/11356/1752},
	copyright    = {Creative Commons - Attribution 4.0 International ({CC} {BY} 4.0)},
	note         = {Slovenian language resource repository {CLARIN}.{SI}}
}

@Inbook{CvRS2023LRS,
author="Krstev, Cvetana and Stankovi{\'{c}}, Ranka",
editor="Rehm, Georg and Way, Andy",
title="Language Report Serbian",
bookTitle="European Language Equality: A Strategic Agenda for Digital Language Equality ",
year="2023",
publisher="Springer International Publishing",
address="Cham",
pages="203--206",
isbn="978-3-031-28819-7",
doi="10.1007/978-3-031-28819-7_32"
}

@inproceedings{stankovic2022distant,
  title={Distant Reading in Digital Humanities: Case Study on the Serbian Part of the ELTeC Collection},
  author={Stankovi{\'c}, Ranka and Krstev, Cvetana and Todorovi{\'c}, Branislava {\v{S}}andrih and Vitas, Du{\v{s}}ko and {\v{S}}kori{\'c}, Mihailo and Ne{\v{s}}i{\'c}, Milica Ikoni{\'c}},
  booktitle={Proceedings of the Thirteenth Language Resources and Evaluation Conference},
  pages={3337--3345},
  year={2022}
}

@inproceedings{stankovic-etal-2020-machine,
    title = "Machine Learning and Deep Neural Network-Based Lemmatization and Morphosyntactic Tagging for {S}erbian",
    author = "Stankovic, Ranka  and
      {\v{S}}andrih, Branislava  and
      Krstev, Cvetana  and
      Utvi{\'c}, Milo{\v{s}}  and
      Skoric, Mihailo",
    editor = "Calzolari, Nicoletta  and
      B{\'e}chet, Fr{\'e}d{\'e}ric  and
      Blache, Philippe  and
      Choukri, Khalid  and
      Cieri, Christopher  and
      Declerck, Thierry  and
      Goggi, Sara  and
      Isahara, Hitoshi  and
      Maegaard, Bente  and
      Mariani, Joseph  and
      Mazo, H{\'e}l{\`e}ne  and
      Moreno, Asuncion  and
      Odijk, Jan  and
      Piperidis, Stelios",
    booktitle = "Proceedings of the Twelfth Language Resources and Evaluation Conference",
    month = may,
    year = "2020",
    address = "Marseille, France",
    publisher = "European Language Resources Association",
    url = "https://aclanthology.org/2020.lrec-1.487",
    pages = "3954--3962",
    language = "English",
    ISBN = "979-10-95546-34-4",
}

@inproceedings{gavrilidou2004building,
  title={Building parallel corpora for eContent professionals},
  author={Gavrilidou, Maria and Labropoulou, Penny and Desipri, Elina and Giouli, Voula and Antonopoulos, Vasilis and Piperidis, Stelios},
  booktitle={Proceedings of the Workshop on Multilingual Linguistic Resources},
  pages={90--93},
  year={2004}
}

@inproceedings{krstev2004multext,
  title={MULTEXT-East resources for Serbian},
  author={Krstev, Cvetana and Vitas, Du{\v{s}}ko and Erjavec, Toma{\v{z}}},
  booktitle={Zbornik 7. mednarodne multikonference Informacijska druzba IS 2004 Jezikovne tehnologije 9-15 Oktober 2004, Ljubljana, Slovenija, 2004},
  year={2004},
  organization={Erjavec, Toma{\v{z}} and Zganec Gros, Jerneja}
}

@article{Miličević_Ljubešić_2016,
title={Tviterasi, tviteraši or twitteraši? Producing and analysing a normalised dataset of Croatian and Serbian tweets}, 
volume={4}, 
url={https://revije.ff.uni-lj.si/slovenscina2/article/view/7007}, 
DOI={10.4312/slo2.0.2016.2.156-188}, 
number={2}, 
journal={Slovenščina 2.0: empirical, applied and interdisciplinary research}, 
author={Miličević, Maja and Ljubešić, Nikola}, 
year={2016}, 
month={9}, 
pages={156–188} }

@inproceedings{samardzic-etal-2017-universal,
    title = "{U}niversal {D}ependencies for {S}erbian in Comparison with {C}roatian and Other {S}lavic Languages",
    author = "Samard{\v{z}}i{\'c}, Tanja  and
      Starovi{\'c}, Mirjana  and
      Agi{\'c}, {\v{Z}}eljko  and
      Ljube{\v{s}}i{\'c}, Nikola",
    booktitle = "Proceedings of the 6th Workshop on {B}alto-{S}lavic Natural Language Processing",
    month = apr,
    year = "2017",
    address = "Valencia, Spain",
    publisher = "Association for Computational Linguistics",
    url = "https://aclanthology.org/W17-1407",
    doi = "10.18653/v1/W17-1407",
    pages = "39--44",
}

@misc{11356/1794,
 title = {Serbian Twitter training corpus {ReLDI}-{NormTagNER}-sr 3.0},
 author = {Ljube{\v s}i{\'c}, Nikola and Erjavec, Toma{\v z} and Batanovi{\'c}, Vuk and Mili{\v c}evi{\'c}, Maja and Samard{\v z}i{\'c}, Tanja},
 url = {http://hdl.handle.net/11356/1794},
 note = {Slovenian language resource repository {CLARIN}.{SI}},
 copyright = {Creative Commons - Attribution-{ShareAlike} 4.0 International ({CC} {BY}-{SA} 4.0)},
 issn = {2820-4042},
 year = {2023} }

@inproceedings{frontini2020named,
  title={Named entity recognition for distant reading in ELTeC},
  author={Frontini, Francesca and Brando, Carmen and Byszuk, Joanna and Galleron, Ioana and Santos, Diana and Stankovi{\'c}, Ranka},
  booktitle={CLARIN Annual Conference 2020},
  year={2020}
}

@inproceedings{rahimi-etal-2019-massively,
    title = "Massively Multilingual Transfer for {NER}",
    author = "Rahimi, Afshin  and
      Li, Yuan  and
      Cohn, Trevor",
    booktitle = "Proceedings of the 57th Annual Meeting of the Association for Computational Linguistics",
    month = jul,
    year = "2019",
    address = "Florence, Italy",
    publisher = "Association for Computational Linguistics",
    url = "https://www.aclweb.org/anthology/P19-1015",
    pages = "151--164",
}

@article{polyglotner,
         author = {Al-Rfou, Rami and Kulkarni, Vivek and Perozzi, Bryan and Skiena, Steven},
         title = {{Polyglot-NER}: Massive Multilingual Named Entity Recognition},
         journal = {{Proceedings of the 2015 {SIAM} International Conference on Data Mining, Vancouver, British Columbia, Canada, April 30- May 2, 2015}},
         month     = {4},
         year      = {2015},
         publisher = {SIAM},
}

@misc{augustyniak2023massively,
      title={Massively Multilingual Corpus of Sentiment Datasets and Multi-faceted Sentiment Classification Benchmark}, 
      author={Łukasz Augustyniak and Szymon Woźniak and Marcin Gruza and Piotr Gramacki and Krzysztof Rajda and Mikołaj Morzy and Tomasz Kajdanowicz},
      year={2023},
      eprint={2306.07902},
      archivePrefix={arXiv},
      primaryClass={cs.CL}
}

@InProceedings{LJUBEI16.340,
  author = {Nikola Ljubešić and Filip Klubička and Željko Agić and Ivo-Pavao Jazbec},
  title = {New Inflectional Lexicons and Training Corpora for Improved Morphosyntactic Annotation of Croatian and Serbian},
  booktitle = {Proceedings of the Tenth International Conference on Language Resources and Evaluation (LREC 2016)},
  year = {2016},
  month = {5},
  location = {Portorož, Slovenia},
  editor = {Nicoletta Calzolari (Conference Chair) and Khalid Choukri and Thierry Declerck and Sara Goggi and Marko Grobelnik and Bente Maegaard and Joseph Mariani and Helene Mazo and Asuncion Moreno and Jan Odijk and Stelios Piperidis},
  publisher = {European Language Resources Association (ELRA)},
  address = {Paris, France},
  isbn = {978-2-9517408-9-1},
  language = {english}
 }

@article{hendrycks2021ethics,
      title={Aligning AI With Shared Human Values},
      author={Dan Hendrycks and Collin Burns and Steven Basart and Andrew Critch and Jerry Li and Dawn Song and Jacob Steinhardt},
      journal={Proceedings of the International Conference on Learning Representations (ICLR)},
      year={2021}
    }

@inproceedings{zellers2019hellaswag,
    title={HellaSwag: Can a Machine Really Finish Your Sentence?},
    author={Zellers, Rowan and Holtzman, Ari and Bisk, Yonatan and Farhadi, Ali and Choi, Yejin},
    booktitle ={Proceedings of the 57th Annual Meeting of the Association for Computational Linguistics},
    year={2019}
}

@misc{tunstall2023zephyr,
      title={Zephyr: Direct Distillation of LM Alignment}, 
      author={Lewis Tunstall and Edward Beeching and Nathan Lambert and Nazneen Rajani and Kashif Rasul and Younes Belkada and Shengyi Huang and Leandro von Werra and Clémentine Fourrier and Nathan Habib and Nathan Sarrazin and Omar Sanseviero and Alexander M. Rush and Thomas Wolf},
      year={2023},
      eprint={2310.16944},
      archivePrefix={arXiv},
      primaryClass={cs.LG}
}

@misc{SlimOrca,
  title = {SlimOrca: An Open Dataset of GPT-4 Augmented FLAN Reasoning Traces, with Verification},
  author = {Wing Lian and Guan Wang and Bleys Goodson and Eugene Pentland and Austin Cook and Chanvichet Vong and "Teknium"},
  year = {2023},
  publisher = {HuggingFace},
  url = {https://https://huggingface.co/Open-Orca/SlimOrca}
}

@misc{cui2023ultrafeedback,
      title={UltraFeedback: Boosting Language Models with High-quality Feedback}, 
      author={Ganqu Cui and Lifan Yuan and Ning Ding and Guanming Yao and Wei Zhu and Yuan Ni and Guotong Xie and Zhiyuan Liu and Maosong Sun},
      year={2023},
      eprint={2310.01377},
      archivePrefix={arXiv},
      primaryClass={cs.CL}
}

@misc{alpaca,
  author = {Rohan Taori and Ishaan Gulrajani and Tianyi Zhang and Yann Dubois and Xuechen Li and Carlos Guestrin and Percy Liang and Tatsunori B. Hashimoto },
  title = {Stanford Alpaca: An Instruction-following LLaMA model},
  year = {2023},
  publisher = {GitHub},
  journal = {GitHub repository},
  howpublished = {\url{https://github.com/tatsu-lab/stanford_alpaca}},
}

@misc{scherrer_yves_2020_3707949,
  author       = {Scherrer, Yves},
  title        = {{TaPaCo: A Corpus of Sentential Paraphrases for 73 Languages}},
  month        = mar,
  year         = 2020,
  publisher    = {Zenodo},
  version      = {1.0},
  doi          = {10.5281/zenodo.3707949},
  url          = {https://doi.org/10.5281/zenodo.3707949}
}

@inproceedings{vitas2008tour,
  title={Tour du monde through the dictionaries},
  author={Vitas, Du{\v{s}}ko and Koeva, Svetla and Krstev, Cvetana and Obradovi{\'c}, Ivan},
  booktitle={Actes du 27eme Colloque International sur le Lexique et la Gammaire},
  pages={249--256},
  year={2008}
}

@inproceedings{tiedemann-2012-parallel,
    title = "Parallel Data, Tools and Interfaces in {OPUS}",
    author = {Tiedemann, J{\"o}rg},
    editor = "Calzolari, Nicoletta  and
      Choukri, Khalid  and
      Declerck, Thierry  and
      Do{\u{g}}an, Mehmet U{\u{g}}ur  and
      Maegaard, Bente  and
      Mariani, Joseph  and
      Moreno, Asuncion  and
      Odijk, Jan  and
      Piperidis, Stelios",
    booktitle = "Proceedings of the Eighth International Conference on Language Resources and Evaluation ({LREC}'12)",
    month = may,
    year = "2012",
    address = "Istanbul, Turkey",
    publisher = "European Language Resources Association (ELRA)",
    url = "http://www.lrec-conf.org/proceedings/lrec2012/pdf/463_Paper.pdf",
    pages = "2214--2218",
}

@inproceedings{hasan-etal-2021-xl,
title = "{XL}-Sum: Large-Scale Multilingual Abstractive Summarization for 44 Languages",
author = "Hasan, Tahmid and
Bhattacharjee, Abhik and
Islam, Md. Saiful and
Mubasshir, Kazi and
Li, Yuan-Fang and
Kang, Yong-Bin and
Rahman, M. Sohel and
Shahriyar, Rifat",
booktitle = "Findings of the Association for Computational Linguistics: ACL-IJCNLP 2021",
month = aug,
year = "2021",
address = "Online",
publisher = "Association for Computational Linguistics",
url = "https://aclanthology.org/2021.findings-acl.413",
pages = "4693--4703",
}

@misc{debruyn2021mfaq,
      title={MFAQ: a Multilingual FAQ Dataset}, 
      author={Maxime {De Bruyn} and Ehsan Lotfi and Jeska Buhmann and Walter Daelemans},
      year={2021},
      eprint={2109.12870},
      archivePrefix={arXiv},
      primaryClass={cs.CL}
}

@inproceedings{hardalov-etal-2020-exams,
    title = "{EXAMS}: A Multi-subject High School Examinations Dataset for Cross-lingual and Multilingual Question Answering",
    author = "Hardalov, Momchil  and
      Mihaylov, Todor  and
      Zlatkova, Dimitrina  and
      Dinkov, Yoan  and
      Koychev, Ivan  and
      Nakov, Preslav",
    editor = "Webber, Bonnie  and
      Cohn, Trevor  and
      He, Yulan  and
      Liu, Yang",
    booktitle = "Proceedings of the 2020 Conference on Empirical Methods in Natural Language Processing (EMNLP)",
    month = nov,
    year = "2020",
    address = "Online",
    publisher = "Association for Computational Linguistics",
    url = "https://aclanthology.org/2020.emnlp-main.438",
    doi = "10.18653/v1/2020.emnlp-main.438",
    pages = "5427--5444",
}

@article{skoric24modeli,
  author    = {Mihailo Škorić},
  title     = {Novi jezički modeli za srpski jezik},
  journal   = {Infoteka},
  volume    = {24},
  issue     = {1},
  year      = {2024},
  publisher = {Zajednica biblioteka univerziteta u Srbiji, Beograd},
  note = {accepted for publishing}
}

@inbook{verbic__2017,
author = {Verbić, Srđan and Suvakov, Milovan and Luzanin, Zorana},
year = {2017},
month = {06},
pages = {},
title = {{NaRDuS} – JAVNOST DOKTORSKIH STUDIJA Transparentnost i otvorenost podataka kroz stvaranje i razvoj nacionalnog repozitorijuma doktorskih disertacija u Srbiji (NaRDuS)},
}

@inproceedings{rayson2000comparing,
  title={Comparing corpora using frequency profiling},
  author={Rayson, Paul and Garside, Roger},
  booktitle={The workshop on comparing corpora},
  pages={1--6},
  year={2000}
}

@phdthesis{pomikalek2011removing,
  title={Removing boilerplate and duplicate content from web corpora},
  author={Pomik{\'a}lek, Jan},
  school={Masaryk university, Faculty of informatics, Brno, Czech Republic},
  year={2011}
}

\label{lastpage}
\end{document}